# Center of gravity PSO for Partitioning Clustering

Shahira Shaaban Azab
*Departmet of computer science, Cairo university*
Shahiraazazy@gmail.com

Hesham Ahmed Hefny
*Departmet of computer science, Cairo university*

*Abstract— This paper presents the local best model of PSO for partition-based clustering. The proposed model gets rid off the drawbacks of gbest PSO for clustering. The model uses a pre-specified number of clusters K. The LPOSC has K neighborhoods. Each neighborhood represents one of the clusters. The goal of the particles in each neighborhood is optimizing the position of the centroid of the cluster. The performance of the proposed algorithms is measured using adjusted rand index. The results is compared with k-means and global best model of PSO.*

## I. INTRODUCTION

Clustering analysis is grouping objects in different clusters such that the objects in the same cluster are similar to each other and different from objects in other clusters. Clustering analysis has many applications in different domains e.g. Biology, Medicine, image processing, pattern recognition, Astronomy, Bioinformatics, Genetics and even Psychiatry[1].

Clustering algorithms can be categorized as hierarchical or partitioning[2]. Hierarchical clustering algorithms group objects in a hierarchical structure. Hierarchal clustering is not sensitive to the initial condition. There is no prior knowledge of the number of clusters. Hierarchal clustering is suitable for categorical data. However, they are computationally expensive, static (object assigned to a cluster cannot be reassigned to another one).

Partitioning clustering algorithms group objects into disjoint clusters. Partitioning clustering problem can be defined as[3]: Let $O$ be a set of $n$ objects.
$O = (o_1, o_2 \dots \dots \dots \dots, o_n)$ where
$O_i = \{o_{i1}, o_{i2}, \dots \dots \dots, o_{id}\} \in \mathbb{R}^d$, $n$ is the number of unlabeled examples, and $d$ number of dimensions. The objective of the clustering algorithm is splitting $n$ unlabeled examples into $K$ clusters Such that:

There is no null cluster.

$$c_i \neq \phi, i = 1, \dots, K \quad (1)$$

In hard clusters, each object belongs to only one cluster.

$$c_i \cap c_j = \phi, i, j = 1, \dots, K \text{ and } i \neq j. \quad (2)$$

In soft clustering, each object may belong to more than one cluster. Objects in fuzzy clustering belong to one or multiple clusters with different membership degree.

The union of all clusters is the original dataset.

$$\cup_{i=1}^{K} c_i = O \quad (3)$$

The next section briefly reviews PSO for optimization. Section 3 summarizes PSO based clustering problem. The Proposed PSO algorithm is illustrated in section 4. Experimental results are summarized in section 5.

## II. BACKGROUND

### A. Particle Swarm Optimization

When a group of birds starts a random search for food in an open area, they actually do not know where the food is. However, they know the best spot found by any member of the flock. Therefore, the best strategy is to follow the nearest bird to the food source taking into consideration the experience of the other birds. If a better spot is found by any member of the swarm, the birds change their direction according to the new information.

PSO mimics the foraging behavior of bird flocks. There is a swarm of particles. The particles represent potential solutions to the problem. Particles fly within the search space trying to find the best points or at least very good ones (like birds searching for food). The swarm cooperatively explores the multidimensional search space. The particles within the swarm exchange information about the successful regions of the search space. The performance of each particle (i.e. the closeness of the particle to the global optimum) is calculated according to a fitness function that varies depending on the optimization problem.

A particle in the swarm changes its positions and velocity according to [4]:

<u>Cognitive components</u>: The best solution found by the particle till now.

<u>Social components</u>: Knowledge of other particles within the swarm; the best position found by any member of the swarm.

### B. PSO Algorithm

Let η be the swarm size, $x_{gbest}$ is the best position found by the swarm, $i$ Current particle index, $x_i, v_i$ are the position and velocity of the current particle respectively, $p_{best}$ represents the best position found by the particle, $g_{best}$ is the index of the global best particle in the entire swarm.

**Initialization**: Population (particles) is generated with random positions $X$ or velocities $V$ or both. The best position vector $P_{best}$ should be different from $X$ in order to make the particles move [4].

**Input**: particles with random positions and velocities.
**Output**: Position of the approximate global optima.
The particles velocity are calculated using (4). The new positions of particles are calculated according to (5)

$$v_i(t+1) = \omega_i * v_i(t) + \varphi_1 * \left(p_{best_i}(t) - x_i(t)\right) \quad (4)$$
$$+ \varphi_2 * (g_{best}(t) - x_i(t))$$

$$x_i(t+1) = x_i(t) + v_i(t) \quad (5)$$

Acceleration coefficients $\varphi_1, \varphi_2$ affect PSO performance significantly. $\varphi_1$ affects the attraction of the particles towards his personal best position, $\varphi_2$ affects the attraction of particles towards the global best particle. $\varphi_1, \varphi_2$ are independently distributed using the uniform distribution in the interval [0,AC].

$$\varphi_1 = rnd_1 * AC_1$$
$$\varphi_2 = rnd_2 * AC_2$$

$rnd_1, rnd_2$ is a random number $\epsilon ]0,1[$

The particle preserves his previous best position $p_{best}$ if $p_{best}$ is better than his current position. However, if his current position is better than $p_{best}$. Then $p_{best}$ changes to the value of the current position. Equation(6) shows update of $p_{best}$ for minimization problems.

$$p_{best_i}(t+1) = \begin{cases} p_{best_i}(t) & \text{if } f(x_i(t+1)) \geq f(p_{best_i}(t)) \\ x_i(t+1) & \text{if } f(x_i(t+1)) < f(p_{best_i}(t)) \end{cases} \quad (6)$$

Global best position $g_{best}$ is the best position achieved by any member of the swarm until the current iteration.

### III. PSO BASED CLUSTERING

PSC can be explained as follows[5]. Given a data set $O$ with $K$ clusters and $D$ attributes. Each particle is represented as a vector of the centroids of classes in the dataset. Hence, each particle is a potential solution to a clustering problem. Thus, the global best particle is the proposed solution for the classification problem. Particles update their positions and velocities to obtain the optimal position for the centroids. Fitness function commonly used to evaluate the performance of particles is the minimum distance between points and potential centroids. In terms of genetic algorithms, the genotype of a particle can be expressed as 2*K*D.

There are $\eta$ particles in the swarm. Each particle $i$ in the swarm is represented by velocities and its positions in the different dimensions at certain time $t$.

Particle position is encoded as $(x_i^1, x_i^2, x_i^3, \ldots \ldots \ldots \ldots x_i^k)_t$.
Particle velocity is encoded as $(v_i^1, v_i^2, v_i^3, \ldots \ldots \ldots \ldots v_i^k)_t$
Where the position of particle $i$ of centroid $j$ in *d dimensions* is
$x_i^j = \{x_{i1}^j, x_{i2}^j, x_{i3}^j, \ldots \ldots \ldots \ldots x_{id}^j\}$
$v_i^j = \{v_{i1}^j, v_{i2}^j, v_{i3}^j, \ldots \ldots \ldots \ldots v_{id}^j\}$
The encoding of position of a single particle $i$ is explained in fig. 1.

**The PSO for clustering algorithm is as follows:**

Initialize the positions and the velocities for each particle in the swarm randomly.
Do while the termination condition is met
    For each particle in population
    For each data point
Calculate distance between data points and all particles of the swarm
Assign data points to the nearest centroid
Calculate the fitness of particles
      IF Current position better than best position
      THEN pbest position= current position.
      IF Current position is better than the global best
      THEN gbest index= current particle index.
      Update particle velocity using (4)
      Move the particle to a new position using (5).
    End For
*End While*

| $x_{i1}^1$ | $x_{i2}^1$ | ... | $x_{id}^1$ | $x_{i1}^2$ | $x_{i2}^2$ | ... | $x_{id}^2$ | ... | $x_{i1}^c$ | $x_{i2}^c$ | ... | $x_{i2}^c$ |
|---|---|---|---|---|---|---|---|---|---|---|---|---|
| $centroid_1$ ||||  $centroid_2$  |||| ... | $centroid_c$ ||||
| Particle $i$ |||||||||||||

Fig. 1. Representation of a particle of global PSO for clustering

Many clustering algorithms based on PSO have been proposed in the literature:

In [5] proposed a PSO cluster algorithm to find the centroid of a user-specified number of clusters and used a hybrid PSO and k-means. K-means is used first to create an initial solution. Then PSO is used to refine centroid which was found by K-means. [6] reported that K-means converge faster. However, PSO has a more accurate solution than k-means.

[7] proposed a dynamic PSO for clustering. They apply their algorithm in the area of image classification. A binary PSO was used to find an optimum number of clusters and k-means refined the class centroid.

[6] proposed image clustering algorithm using PSO. Each cluster group similar pixels together and PSO is optimizing the centroids of a predefined number of clusters. This proposed algorithm applied to different types of images: synthesized image, an MRI image of the human brain, and a satellite image of Lake Tahoe. Compared to the state of the art algorithm ISODATA, it showed promising results. PSO is used to optimize the centroids of clusters and choose the optimum number of clusters. Results assured that the state of the art algorithm convergences faster than PSO but with inaccurate clustering.

[8] proposed two algorithms using PSO for data clustering. The first hybrid algorithm started with running K-means algorithm for 1000 iteration or threshold of 0.0001 then the output of the K-means algorithm fed as a particle to the PSO and other particles are initialized randomly.

## IV. THE PROPOSED ALGORITHM

The idea of LCPSO based on LDWMean PSO presented in[9]. The velocity equation of LCPSO, see equation7 uses the lbest model instead of gbest model presented by the author in [9]. The proposed LCPSO accelerate the convergence of the algorithm towards the centroid of the cluster. In the LCPSO algorithm, the center of the cluster works as the center of gravity attracts particles.

$$v_i(t+1) = \omega_i * v_i(t) + \varphi_1 * \left(\frac{p_{best}+l_{best}}{2} - x_i(t)\right) + \varphi_2 * \left(\frac{p_{best}-l_{best}}{2} - x_i(t)\right) \quad (7)$$

The proposed algorithm can be summarized as follows:
Input: unlabeled dataset
distance function d() usually Euclidian Distance
number of cluster(number of neighbors)
test instance
Assign particles to neighborhoods almost equally
Initialize particle position and velocity randomly
Assign swarm_fitness to inifinity
Do while the termination condition is met
   For each neighborhood
    For each particle in neighborhood
    For each data point
Calculate distance between data points and all particles of the swarm
Assign data points to the nearest particle
Calculate the fitness of particles(mean Euclidian distance+labels data)
Assign data points to the neighborhoods of the associated particles
     IF Current position better than best position
     THEN pbest position= current position.
     IF Current position is better than the neighborhood best
     THEN lbest index= current particle index.
   Update particle velocity using (7)
   Move the particle to a new position using (5).

## V. EXPERIMENTS AND RESULTS

The proposed algorithms are tested using four synthetic dataset and twelve real-world datasets. The datasets are selected from different domains with a different number of instances and attributes, see Table I A. The results are compared with the state-of-the-art PSO based Clustering algorithm and K-means algorithm, LPSO[10].

TABLE I PROPERTIES OF ARTIFICIAL DATASETS

| Datasets | Features | Instances | Classes |
|---|---|---|---|
| banana | 2 | 5300 | 2 |
| aggregation | 2 | 788 | 7 |
| compound | 2 | 399 | 6 |
| two-norm | 20 | 7400 | 2 |

TABLE II PROPERTIES OF TWO-CLASS DATASETS

| Datasets | Features | Instances | Classes |
|---|---|---|---|
| haberman | 3 | 306 | 2 |
| titanic | 3 | 2201 | 2 |
| pima | 8 | 768 | 2 |
| wisconsin | 10 | 699 | 2 |

TABLE III SIMULATION RESULTS FOR MULTI-CLASS DATASETS

| Datasets | Features | Instances | Classes |
|---|---|---|---|
| Iris | 4 | 150 | 3 |
| Abalone | 8 | 4177 | 28 |
| Pen | 16 | 10992 | 10 |
| SatImage | 36 | 6435 | 7 |

TABLE IV SIMULATION RESULTS FOR HIGH-DIMENSION DATASETS

| Datasets | Features | Instances | Classes |
|---|---|---|---|
| ups | 1500 | 241 | 2 |
| coil | 1500 | 241 | 6 |
| bci | 400 | 117 | 2 |
| isolet | 7797 | 617 | 26 |

TABLE V SIMULATION RESULTS FOR ARTIFICIAL DATASETS

| Datasets | Adjusted Rand Index | | | |
|---|---|---|---|---|
| | K-means | PSOC | LPSO | LCPSO |
| banana | 0.3923 | 0.3934 | 0.5681 | 0.5981 |
| aggregation | 0.4723 | 0.4901 | 0.5188 | 0.5188 |
| compound | 0.4295 | 0.2955 | 0.4696 | 0.4996 |
| two-norm | 0.1551 | 0.156 | 0.1601 | 0.1701 |

TABLE VI SIMULATION RESULTS FOR TWO-CLASS DATASETS

| Datasets | Adjusted Rand Index | | | |
|---|---|---|---|---|
| | K-means | PSOC | LPSO | LCPSO |
| haberman | 0.3869 | 0.4663 | 0.707 | 0.7970 |
| titanic | 0.6574 | 0.3918 | 0.7733 | 0.8733 |
| pima | 0.2611 | 0.2796 | 0.303 | 0.4830 |
| wisconsin | 0.5919 | 0.5982 | 0.616 | 0.6760 |

TABLE VII SIMULATION RESULTS FOR MULTI-CLASS DATASETS

| Datasets | Adjusted Rand Index | | | |
|---|---|---|---|---|
| | K-means | PSOC | LPSO | LCPSO |
| iris | 0.5043 | 0.5421 | 0.552 | 0.6320 |
| abalone | 0.3122 | 0.249 | 0.1641 | 0.3144 |
| pen | 0.3194 | 0.366 | 0.4307 | 0.5807 |
| satimage | 0.3552 | 0.226 | 0.3664 | 0.3864 |

TABLE VIII SIMULATION RESULTS FOR HIGH-DIMENSIONAL DATASETS

| Datasets | Adjusted Rand Index | | | |
|---|---|---|---|---|
| | K-means | PSOC | LPSO | LCPSO |
| ups | 0.0471 | 0.0173 | 0.0488 | 0.0788 |
| coil | 0.0038 | 0.0408 | 0.0417 | 0.0717 |
| BCI | 0.2001 | 0.0959 | 0.2018 | 0.3218 |
| isolet | 0.0001 | 0.0362 | 0.0593 | 0.0593 |

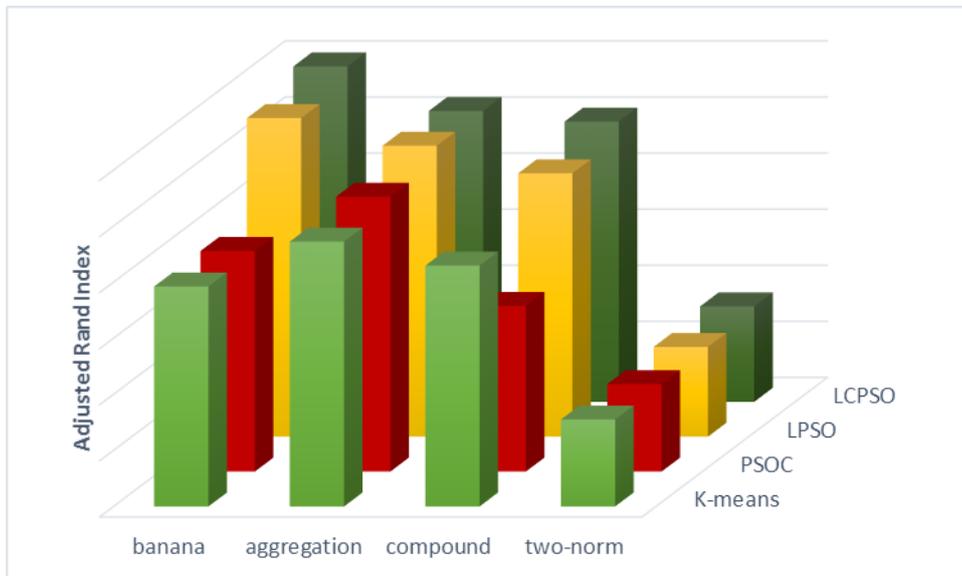

Fig. 2. Adjusted RI for Artificial Datasets

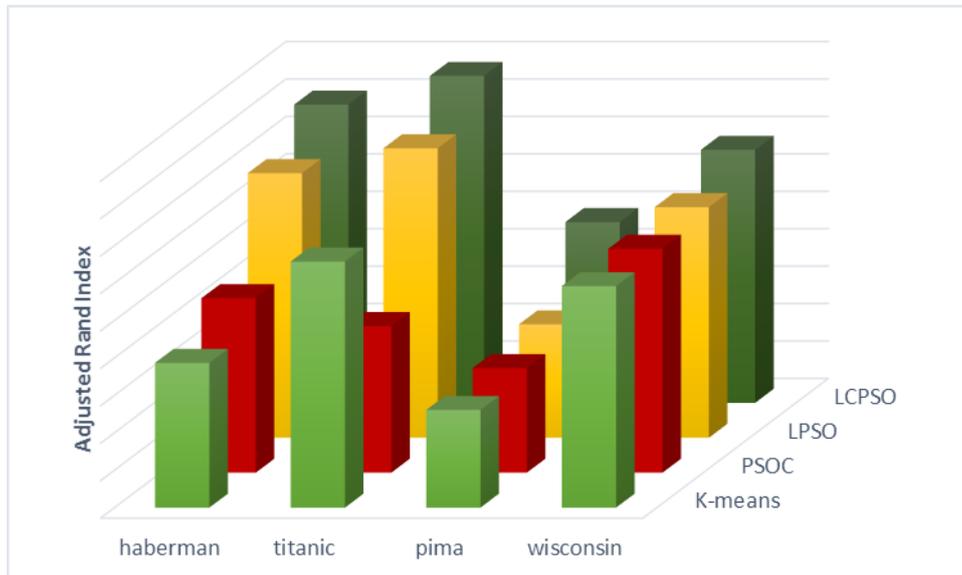

Fig. 3. Adjusted RI for Two-Class Datasets

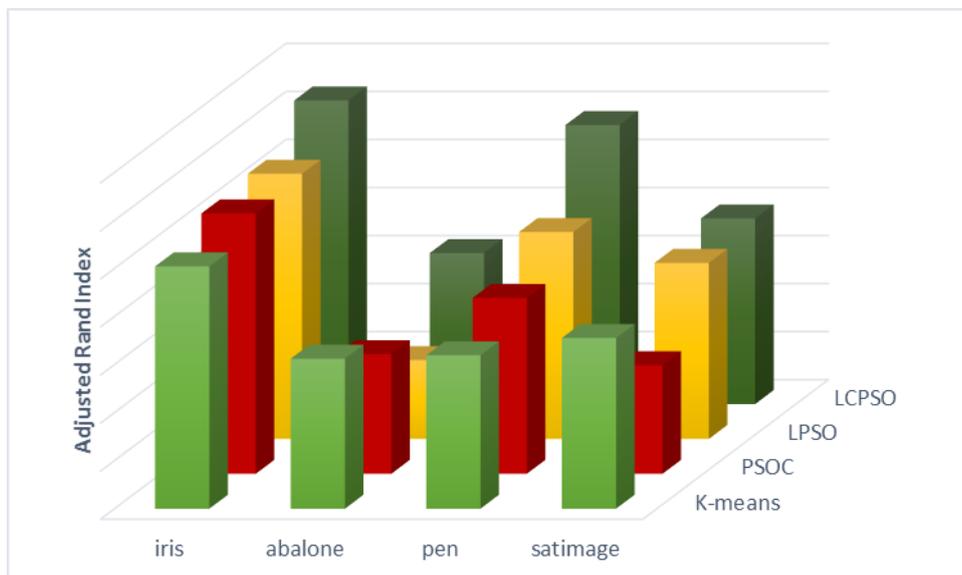

Fig. 4. Adjusted RI for Multi-Class Datasets

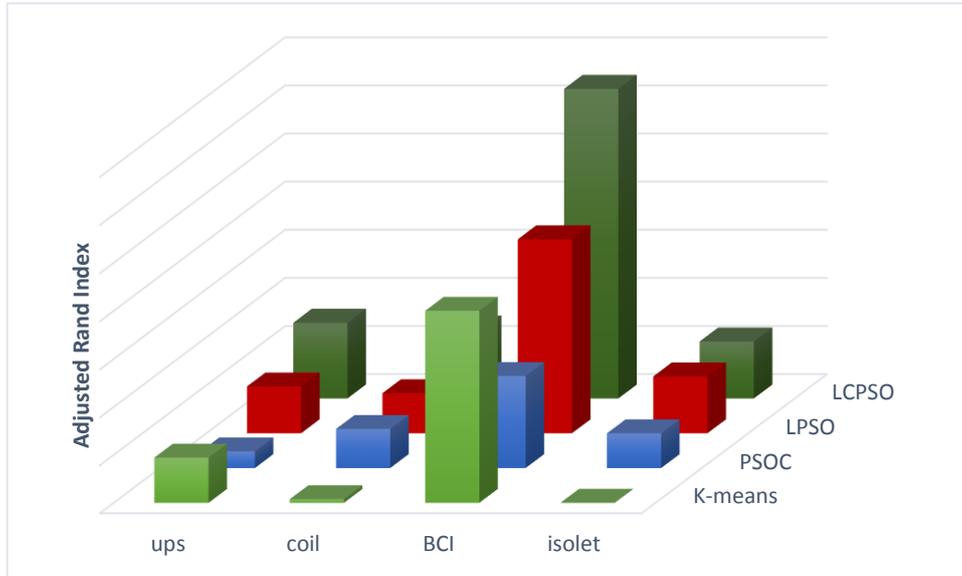

Fig. 5. Adjusted RI for High-Dimension Datasets

The performance of the proposed algorithms is measured using adjusted rand index. Since K-means and PSO-based clustering assume a separation between clusters, and they adopts the distance to measure the similarity between objects, there is an assumption that similar objects are close and belongs to the same clusters. The number of classes affects the performance of the proposed model for instance for the data set satimage with 36 class performance of LCPSO is 38% compared to k-means 52%.

The analysis of the experiments shows that number of particles affects the performance of the algorithm. With 200 particles the performance of LCPSO is about %38, see figure 6. The performance of the LCPSO with high dimension dataset is poor. Increasing number of particles has and insignificant effect on the performance of PSO with high dimension dataset. The Number of particles is a parameters setting that has significant effects on the performance of PSO. Choosing the appropriate number of particles for specific dataset needs more analysis. It has a relationship with the number of clusters in the dataset. This relationship will be investigated as a future work.

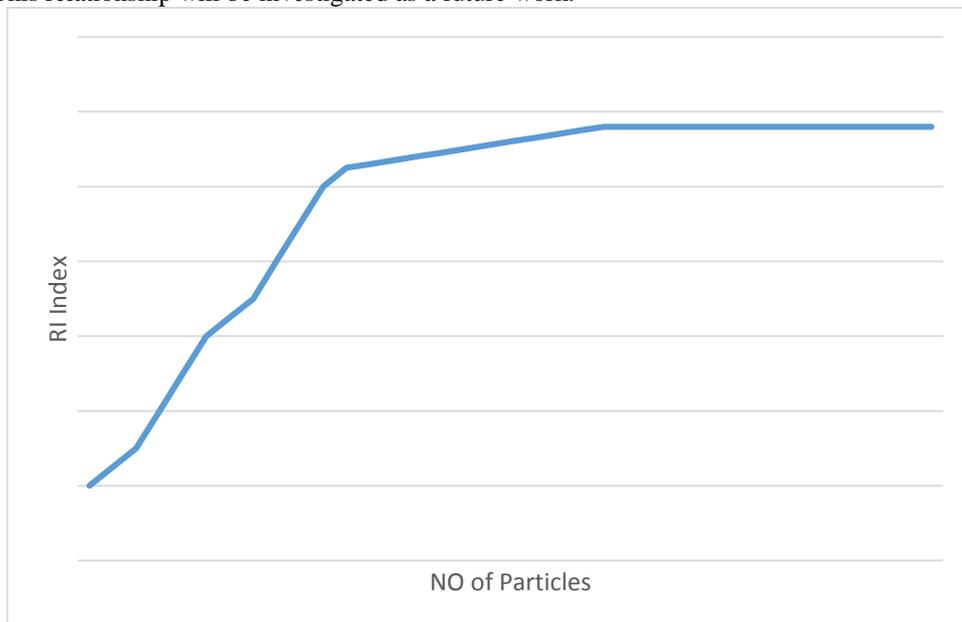

Fig 6. Rand Index VS number of Particles for satimage Dataset

VI. CONCLUSION

This Paper presents a local version of PSO for clustering. The global version of PSO is a special case of the local version where the size of the neighborhood is the whole swarm. The convergence of local PSO version is slower than global version. However, it has a better chance of finding the global optimum. Though, the gbest model converges faster than lbest. It can be trapped easily in a local optimum since all individuals are attracted to the already found solution. While lbest version particle attracted towards different best particles leads to more diversity, so this version is more resistant to be trapped in local minima. The Experiments assures the superiority of the algorithm.